\documentclass[10pt,twocolumn,letterpaper]{article}

\usepackage{cvpr}
\usepackage{times}
\usepackage{epsfig}
\usepackage{graphicx}
\usepackage{amsmath}
\usepackage{amssymb}

\usepackage{setspace}
\usepackage{booktabs}

\usepackage[breaklinks=true,bookmarks=false]{hyperref}

\cvprfinalcopy 

\def\cvprPaperID{4789} 

\begin{document}

\title{Attention Based Glaucoma Detection: \\A Large-scale Database and CNN Model}

\author{Liu~Li\textsuperscript{$\dagger$},~
        Mai~Xu\textsuperscript{$\dagger$$\ddagger$}\thanks{Mai Xu is the corresponding author of this paper.}~,~
        Xiaofei~Wang\textsuperscript{$\dagger$},~
        Lai~Jiang\textsuperscript{$\dagger$},~
        Hanruo~Liu\textsuperscript{$\S$},~\\
$\dagger$ School of Electronic and Information Engineering, Beihang University, Beijing, China \\
$\ddagger$ Hangzhou Innovation Institute，Beihang University, Hangzhou, Zhejiang, China \\
$\S$Beijing Institute of Ophthalmology, Beijing Tongren Hospital, Beijing, China\\
{\tt\small $\dagger$\{liliu1995, maixu, xfwang, jianglai.china\}@buaa.edu.cn}
}

\maketitle
\thispagestyle{empty}
\pagestyle{empty}
\begin{abstract}

Recently, the attention mechanism has been successfully applied in convolutional neural networks (CNNs), significantly boosting the performance of many computer vision tasks.
Unfortunately, few medical image recognition approaches incorporate the attention mechanism in the CNNs.
In particular, there exists high redundancy in fundus images for glaucoma detection, such that the attention mechanism has potential in improving the performance of CNN-based glaucoma detection.
This paper proposes an attention-based CNN for glaucoma detection (AG-CNN).
Specifically, we first establish a large-scale attention based glaucoma (LAG) database, which includes 5,824 fundus images labeled with either positive glaucoma (2,392) or negative glaucoma (3,432).
The attention maps of the ophthalmologists are also collected in LAG database through a simulated eye-tracking experiment.
Then, a new structure of AG-CNN is designed, including an attention prediction subnet, a pathological area localization subnet and a glaucoma classification subnet.
Different from other attention-based CNN methods, the features are also visualized as the localized pathological area, which can advance the performance of glaucoma detection.
Finally, the experiment results show that the proposed AG-CNN approach significantly advances state-of-the-art glaucoma detection.

\end{abstract}


\section{Introduction}

\begin{figure}[h]

\centering
\includegraphics[width=.77\linewidth]{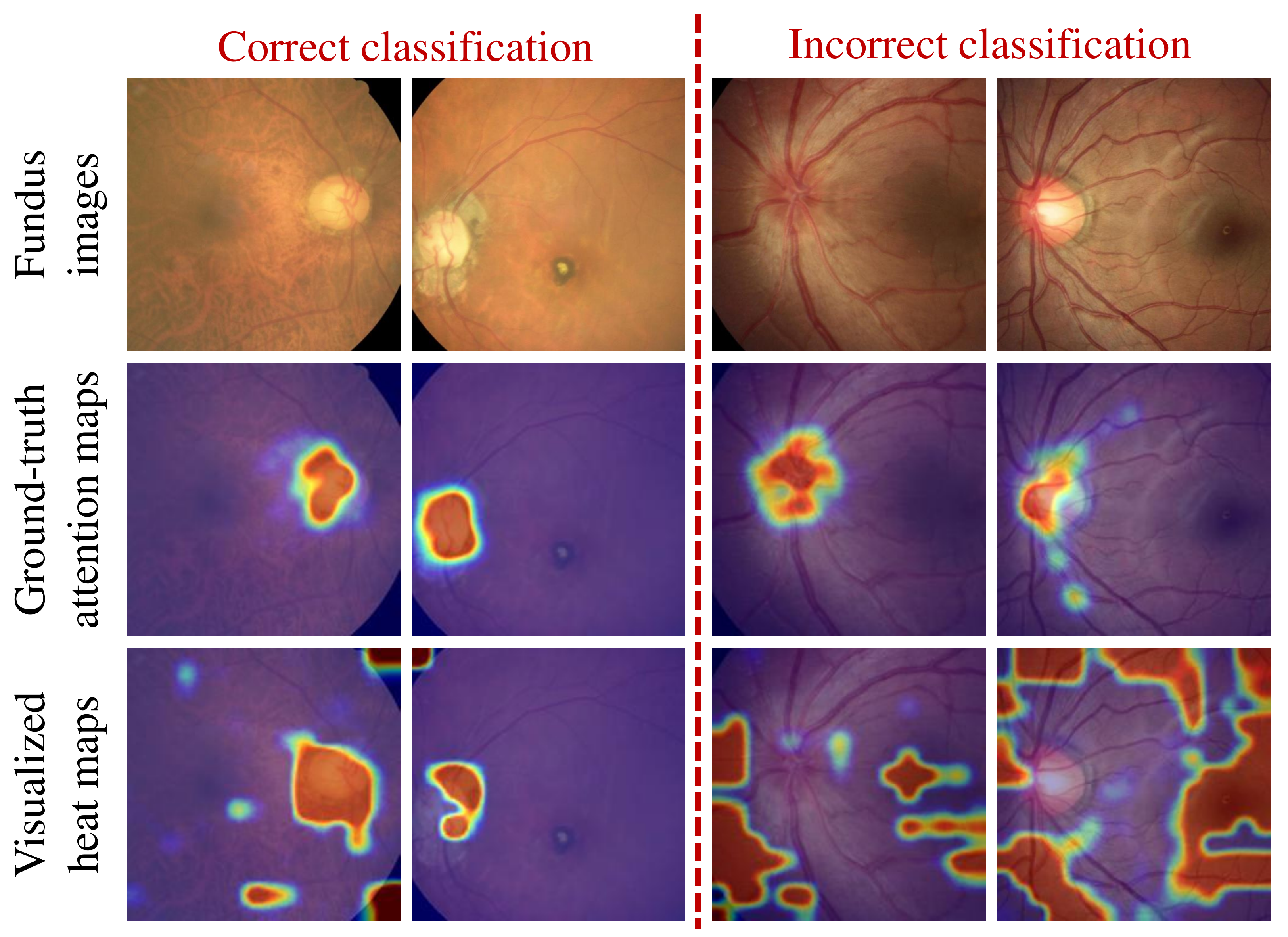}

\caption{\footnotesize Examples of glaucoma fundus images, attention maps by ophthalmologists in glaucoma diagnosis and visualization results of a CNN model (Bottom) \cite{he2016deep} by an occlusion experiment \cite{Matthew2013Visualizing}. 
The Pearson Correlation Coefficient (CC) results between the visualized heat maps and ground-truth ophthalmologist attention maps are 0.33 and 0.14 for correct and incorrect glaucoma classification, respectively.
}
\label{fig1}

\end{figure}

In recently years, the attention mechanism has been successfully applied in deep learning based computer vision tasks, i.e., object detection \cite{ba2015multiple,simonyan2014very,Ren2015Faster}, image caption \cite{xu2015show,yu2017supervising,anderson2018bottom} and action recognition \cite{sharma2015action}. The basic idea of the attention mechanism is to locate the most salient parts of the features in deep neural networks (DNNs), such that redundancy is removed for the vision tasks. In general, the attention mechanism is embedded in DNNs by leveraging the attention maps. Specifically, on the one hand, the attention maps in \cite{simonyan2014very,Ren2015Faster,xu2015show,sharma2015action} are yielded in a self-learned pattern, with other information weakly supervising the attention maps, i.e., the classification labels. On the other hand, \cite{yu2017supervising,xu2017subjective} utilize the human attention information to guide the DNNs focusing on the region of interest (ROI).

Redundancy also exists in medical image recognition, interfering the recognition results.
In particular, there exists heavy redundancy in fundus images for disease recognition.
For example, the pathological areas of fundus images are in the region of optic cup and disc, or its surrounding blood vessel and optic nerve area \cite{Liang1997Supernormal}; other regions such as the boundary of the eye ball are redundant for the medical diagnosis.
As shown in Figure \ref{fig1}, glaucoma, an irreversible optic disease, can be correctly detected by a convolutional neural network (CNN) \cite{he2016deep}, when the visualized heat maps are consistent with the attention maps of ophthalmologists.
Otherwise, glaucoma is mislabeled by the CNN model when the visualized heat maps focus on redundant regions.
Therefore, it is reasonable to combine the attention mechanism in the CNN model for using fundus images to detect ophthalmic disease.

However, to our best knowledge, there has been no works incorporating the human attention in medical image recognition. This is mainly because there lacks the doctor attention database, which needs the qualified doctors and a special technique of capturing the doctor attention in the diagnosis.
As such, in this paper, we first collect a large-scale attention based fundus image database for glaucoma detection (LAG), including 5,824 images with diagnose labels and human attention maps. 
Based on the real human attention,
we propose an attention based CNN method (called AG-CNN) for glaucoma detection based on fundus images.

Although human attention is able to reduce heavy redundancy in fundus images for disease recognition, it may also miss some of the pathological area which is helpful for disease detection. As a result,
the existing CNN models have outperformed the doctors in medical image recognition \cite{Kermany2018Identifying,Rajpurkar2017CheXNet,Poplin2017Predicting}.
Thus, we propose to refine the predicted attention maps by incorporating a feature visualization structure for glaucoma detection. As such, the gap between human attention and pathological area can be bridged.
In fact, there have been several methods for automatically locating the pathological area \cite{zhang2017weakly, gondal2017weakly,feng2017discriminative,ge2017skin,li2017thoracic}, based on the class activation mapping model (CAM) \cite{zhou2015learning}.
However, these methods cannot locate the pathological area at a small region due to the limitation of its feature size.
In this paper, we employ the guided back propagation (BP) method \cite{springenberg2014striving} to locate the tiny pathological area, based on the predicted attention maps.  Consequently, the attention maps can be refined and then used to highlight the most critical regions for glaucoma detection.

The main contributions of this paper are: (1) We establish a LAG database with 5,824 fundus images, along with their labels and attention maps.
(2) We propose incorporating the attention maps in AG-CNN, such that the  redundancy can be removed from fundus images for glaucoma detection.
(3) We develop a new architecture of AG-CNN, which visualizes the CNN feature maps for locating pathological area and then classifies binary glaucoma.


\section{Medical Background}

The recent success of deep learning methods has benefitted medical diagnosis \cite{Esteva2017Dermatologist,Chen2016Mitosis,Yu2017Volumetric}, especially for automatically detecting oculopathy in fundus images \cite{Gulshan2016Development,Gargeya2017Automated,Ting2017Development}. Specifically, \cite{Gulshan2016Development,Gargeya2017Automated} worked on classification of diabetic retinopathy using the CNN models. \cite{Ting2017Development} further proposed deep learning systems for multi-ophthalmological diseases detection. However, the above works all transfered some classic CNN model for nature image classification to medical image classification, regardless of the characteristic of fundus images.

Glaucoma detection methods can be basically divided into 2 categories, i.e., heuristic methods and deep learning methods. The heuristic glaucoma detection methods extract features based on some image processing techniques \cite{Acharya2011Automated,Dua2012Wavelet,Issac2015An,Singh2016Image}. Specifically, \cite{Acharya2011Automated} extracted the texture features and higher order spectra features for glaucoma detection. \cite{Dua2012Wavelet} used the wavelet-based energy features for glaucoma detection. Both \cite{Acharya2011Automated,Dua2012Wavelet} applied support vector machine (SVM)
and naive Bayesian classifier to classify the hand-crafted features. However, the above heuristic methods only consider a handful of features on fundus images, leading to lower classification accuracy.

Another category of glaucoma detection methods is based on deep learning \cite{Shankaranarayana2017Joint,Zilly2017Glaucoma,Chen2015Glaucoma,Li2016Integrating,Z2018Efficacy}. Specifically, \cite{Shankaranarayana2017Joint,Zilly2017Glaucoma} reported their deep learning work on glaucoma detection based on automatic segmentation of optic cup and disc. However, their work assume that only the optical cup and disc are related to glaucoma, lacking end-to-end training. On the other hand, \cite{Chen2015Glaucoma} firstly proposed a CNN method for glaucoma detection in an end-to-end mannar. \cite{Li2016Integrating} followed Chen's work and proposed an advanced CNN structure combining the holistic and local features for glaucoma classification. To regularize the input images, both \cite{Chen2015Glaucoma,Li2016Integrating} preprocessed the original fundus images to remove the redundant regions. However, due to the limited training data and simple structure of networks, the previous works did not achieve high sensitivity and specificity. Most recently,  a deeper CNN structure has been proposed in \cite{Z2018Efficacy}. However, the fundus images exist large redundancy irrelevant to glaucoma detection, leading to the low efficiency for \cite{Z2018Efficacy}.

\section{Database}

\subsection{Establishment}

In this work, we establish a large-scale attention based glaucoma detection database. Our LAG database contains 5,824 fundus images with 2,392 positive and 3,432 negative glaucoma samples obtained from Beijing Tongren Hospital \footnote{The database is available at https://github.com/smilell/AG-CNN.}. Our work is conducted according to the tenets of Helsinki Declaration. As the retrospective nature and fully anonymized usage of color retinal fundus images, we are exempted by the medical ethics committee to inform the patients. Each fundus image is diagnosed by qualified glaucoma specialists, taking the consideration of both morphologic and functional analysis, i.e, intra-ocular pressure, visual field loss and manual optic disc assessment. As a result, the binary labels of positive or negative glaucoma of all fundus images are confirmed, seen as the gold standard.

\begin{figure}

\centering
\includegraphics[width=1.\linewidth]{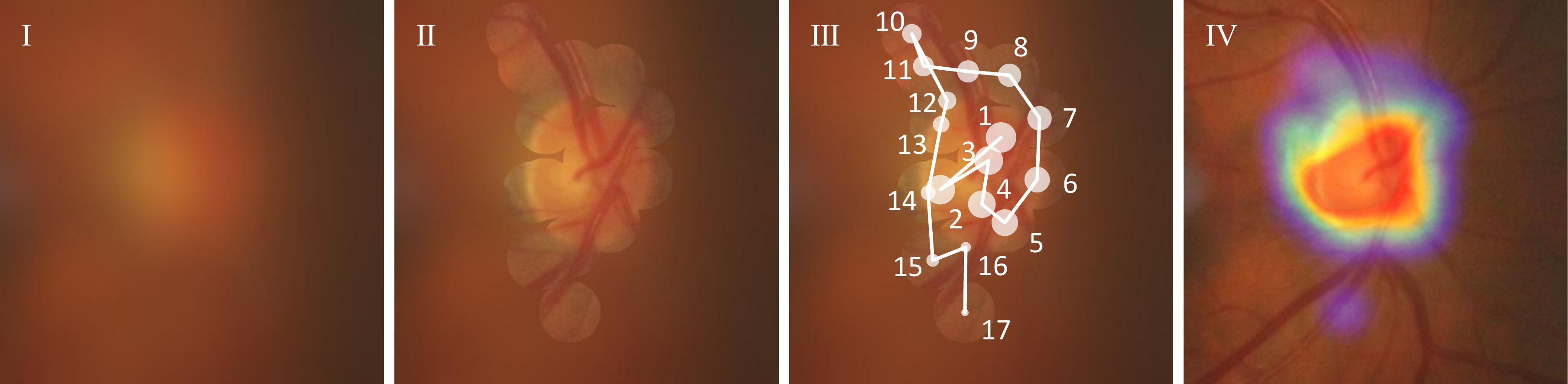}

\caption{\footnotesize An example of capturing fixations of an ophthalmologist in glaucoma diagnosis. (Left): Original blurred fundus images. (Middle-left): Fixations of the ophthalmologist with cleared regions. (Middle-right): The order of clearing the blurred regions. Note that the size of the white circles represents the order of fixations. (Right): The generated attention map based on the captured fixations. }

\label{fig:2}
\end{figure}

\begin{table}

\scriptsize
\centering
\begin{spacing}{1.1}
\caption{\footnotesize CC values of attention maps between one ophthalmologist and the mean of the rest ophthalmologists.}
\label{table:1}

\begin{tabular}{ccc}
    \toprule
      Ophthalmologist & one v.s. others & one v.s. random \\
    \midrule
      $1^{st}$ & 0.594 & $6.59\times10^{-4}$\\
      $2^{nd}$ & 0.636 & $2.49\times10^{-4}$ \\
      $3^{rd}$ & 0.687 & $2.49\times10^{-4}$ \\
      $4^{th}$ & 0.585 & $8.44\times10^{-4}$ \\
    \bottomrule

\end{tabular}
\end{spacing}

\end{table}
Based on the above labelled fundus images, we further conduct an experiment to capture the attention regions of the ophthalmologists in glaucoma diagnosis. The experiment is based on an alternative method for eye tracking \cite{Kim2017BubbleView}, in which mouse clicks are used by the ophthalmologists to explore ROI for glaucoma diagnosis. Specifically, all the fundus images are initially displayed blurred, and then the ophthalmologists use the mouse as an eraser to successively clear the circle regions for diagnosing glaucoma.
Note that the radius of all circle regions is set to $40$ pixels, while all fundus images are with $500 \times 500$ pixels.
This ensures that the circle regions are approximately equivalent to the fovea ($2^\circ-3^\circ$) of the human vision system at a comfortable viewing distance (3-4 times of screen height).
The order of clearing the blurred regions represents the degree of attention by ophthalmologists, as the GT of the attention map.
Once the ophthalmologist is able to diagnose glaucoma with the partly cleared fundus image, the above region clearing process is terminated and the next fundus image is displayed for diagnosis.

In the above experiment, the fixations of ophthalmologists are represented by the center coordinate $(x_i^j,y_i^j)$ of the cleared circle region for the $i$-th fixation of the $j$-th ophthalmologist. Then, the attention map $\mathbf{A}$ of one fundus image can be generated by convoluting all fixations $\{(x_i^j,y_i^j)\}_{i=1,j=1}^{I_j,J}$ with the 2D Gaussian filter at square decay according to the order of $i$, where $J$ is the total number of ophthalmologists (=4 in our experiment) and $I_j$ is the number of fixations from the $j$-th ophthalmologist on the fundus image. Here, the standard deviation of the Gaussian filter is set to 25, according to \cite{xu2017learning}. Figure \ref{fig:2} shows an example of the fixations of one ophthalmologist and attention map of all ophthalmologists for a fundus image.

\begin{figure}[!t]

\centering
\includegraphics[width=1.\linewidth]{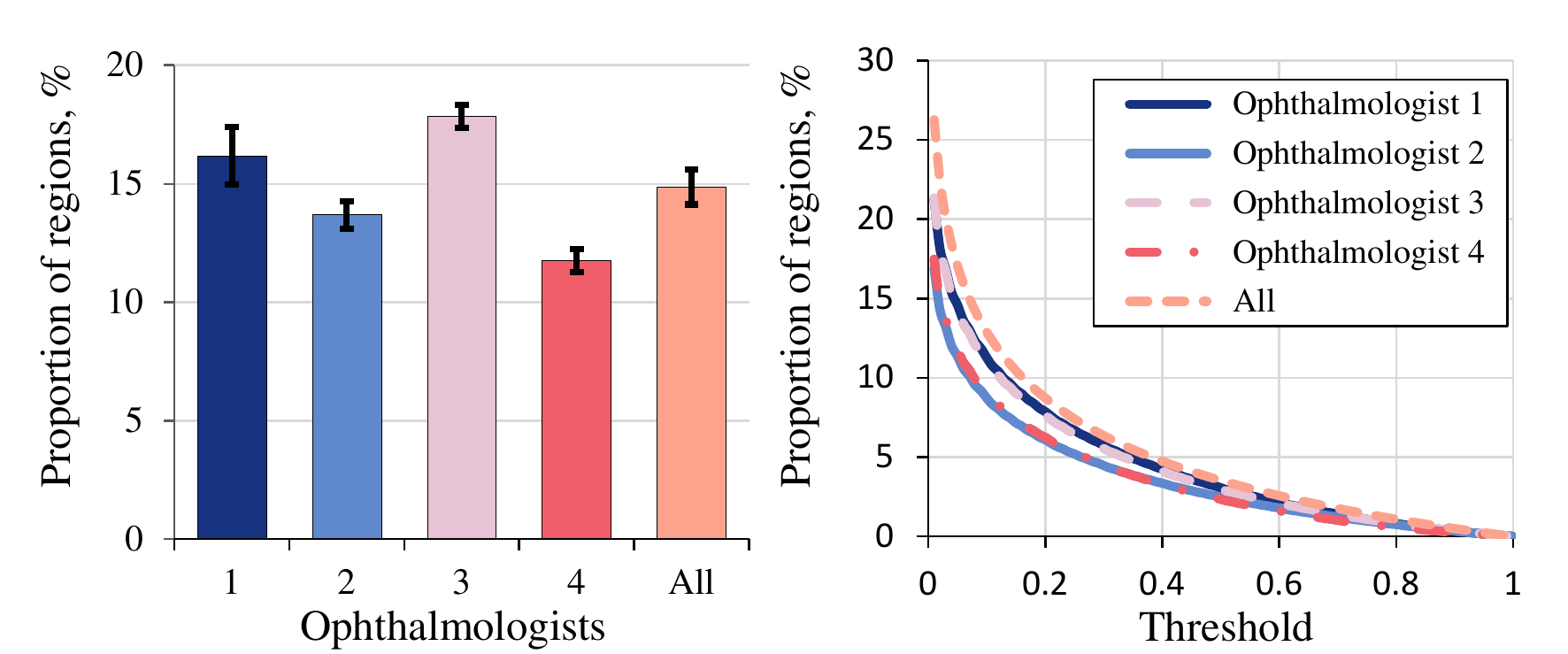}
\caption{\footnotesize (Left): Proportion of regions in the fundus images cleared by different ophthalmologists for glaucoma diagnosis. (Right): Proportion of regions in attention maps with values being above a varying threshold. Note that the values of the attention maps range from 0 to 1.}
\label{fig:4}

\end{figure}

\subsection{Data analysis}

Now, we mine our LAG database to investigate the attention maps of all fundus images in glaucoma diagnosis. Specifically, we have the following findings.

	\textit{Finding 1}: \textit{The ROI in fundus images is consistent across ophthalmologists for glaucoma diagnosis.}

\textit{Analysis}: In this analysis, we calculate the Pearson correlation coefficients (CC) of attention maps between one ophthalmologist and the remaining three ophthalmologists. Table \ref{table:1} reports the CC results averaged over all fundus images in our LAG database. In this table, we also show the CC results of attention maps between one ophthalmologist and the random baseline. Note that the random baseline generates the attention maps by making their values follow the Gaussian distribution. We can see from Table \ref{table:1} that the CC values of attention maps between one and the remaining ophthalmologists are all above 0.55, significantly larger than those of the random baseline. This implies that attention exists consistency among ophthalmologists in glaucoma diagnosis. This completes the analysis of \textit{Finding 2}.

	\textit{Finding 2}: \textit{The ROI in fundus images concentrates on small regions for glaucoma diagnosis.}
	
	\textit{Analysis}: In this analysis, we calculate the percentage of regions that ophthalmologists cleared for glaucoma diagnosis. Figure \ref{fig:4} (Left) shows the percentage of the cleared circle regions for each ophthalmologist, which is averaged over all 5,824 fundus images of our LAG database. We can see that the average ROI accounts for 14.3\% of the total area in the fundus images, with a maximum of 17.8\% (the $3^{rd}$ ophthalmologist) and a minimum of 11.8\% (the $4^{th}$ ophthalmologist). Moreover, we calculate the proportion of regions in attention maps, the values of which are above a varying threshold. The result is shown in Figure \ref{fig:4} (Right). The fast decreasing curve shows that most attention only focuses on small regions of fundus images for  glaucoma diagnosis. This completes the analysis of \textit{Finding 2}.

\begin{figure}[!t]

\centering
\includegraphics[width=.75\linewidth]{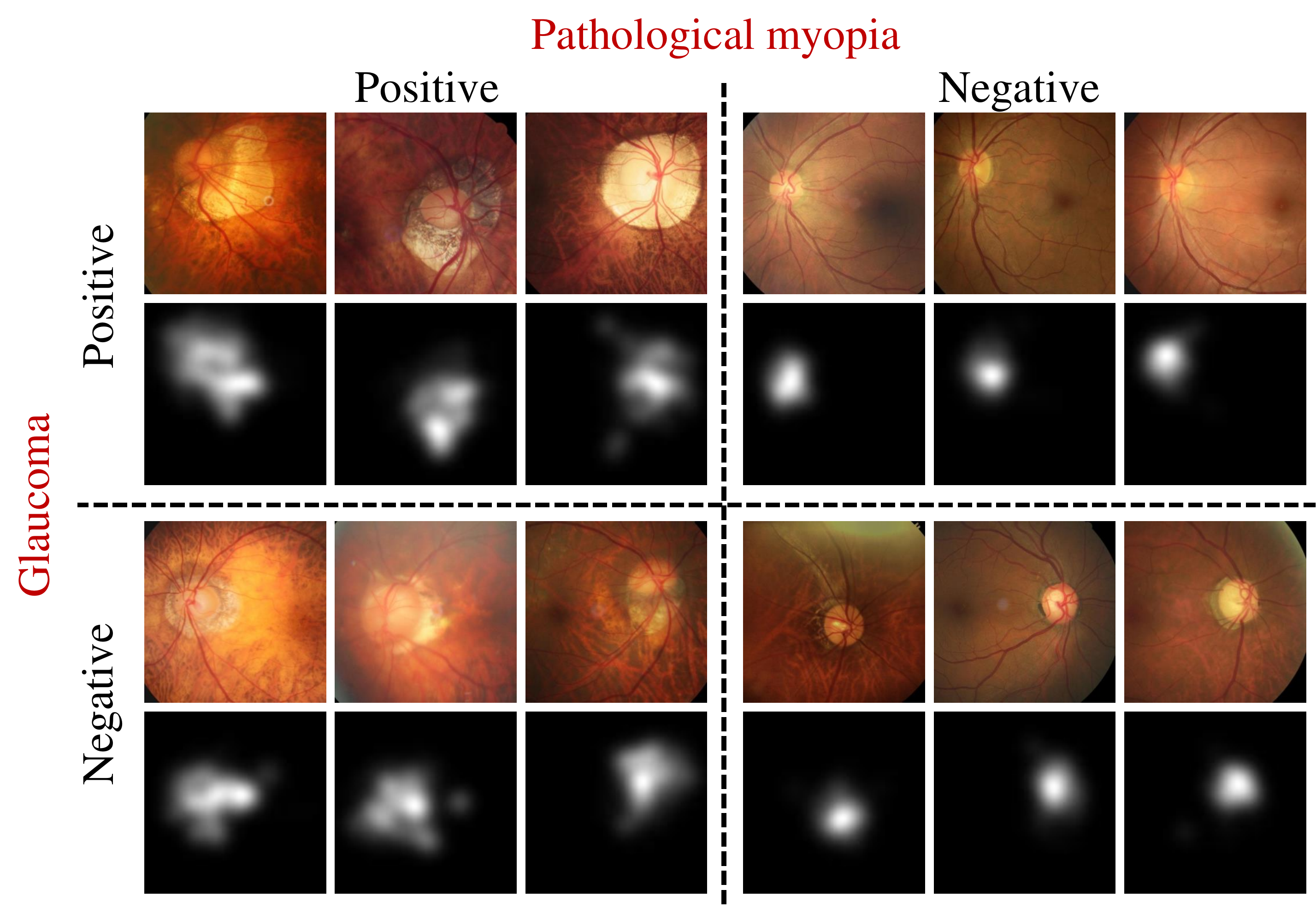}
\caption{\footnotesize Fundus images with or without glaucoma for both positive and negative pathological myopia.
}

\label{scale}

\end{figure}
	
\textit{Finding 3}: \textit{The ROI for glaucoma diagnosis is of different scales.}

\begin{figure}[!t]

\centering
\includegraphics[width=.7\linewidth]{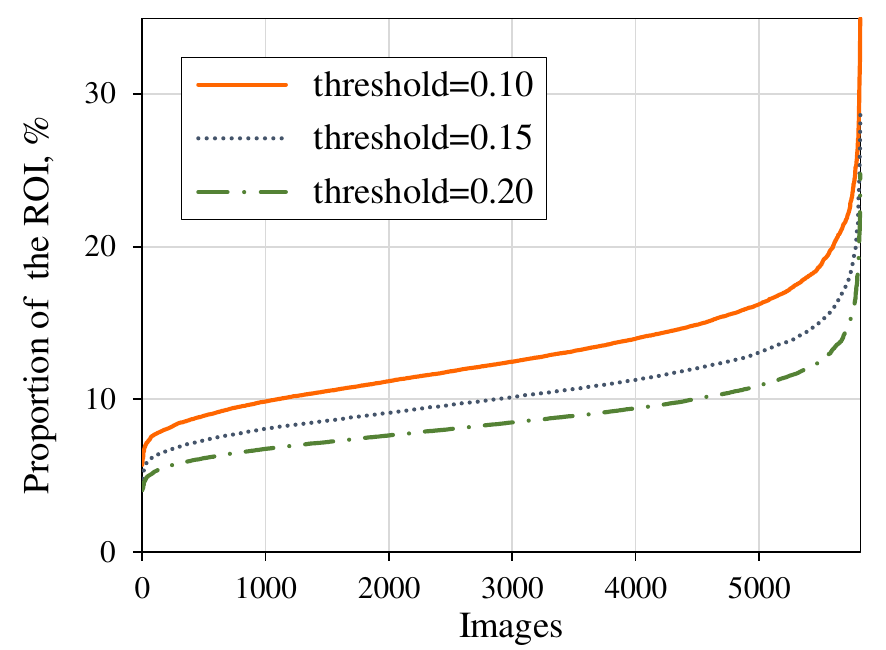}

\caption{\footnotesize Proportion of ROI above the threshold of 0.10, 0.15 and 0.20, for all of the fundus images in LAG database.
}

\label{region3}

\end{figure}

    \textit{Analysis}: \textit{Finding 2} shows that the ROI is small for glaucoma diagnosis, comparing with the whole fundus images. Here, although ROI is small, its scale is various across all the fundus images. Figure \ref{scale} visualizes the fixation maps of some fundus images, in which the ROI are with different scales.
    As shown in Figure \ref{scale}, the sizes of the optic discs for pathological myopia are considerably larger than others.
    As such, we use myopia and non-myopia to select samples with various scales of ROI (large or small optic cups). We further find that the images of both positive and negative glaucoma have various-scaled ROI, as demonstrated in Figure \ref{scale}.
    For each image in our LAG database, Figure \ref{region3} further plots the proportion of the ROI in the fixation maps, the values of which are larger than a threshold. We can see that the ROI is at different scale for glaucoma diagnosis. Finally, the analysis of \textit{Finding 3} can be accomplished.



\section{Method}

\subsection{Framework}

In this section, we discuss the proposed AG-CNN method.
Since \textit{Findings 1} and \textit{2} show that glaucoma diagnosis is highly related to small ROI regions,
the attention prediction subnet is developed in AG-CNN for reducing the redundancy of fundus images.
In addition, we design a pathological area localization subnet, which is achieved by visualizing the CNN feature map, based on ROI regions of the attention prediction subnet.
Based on the pathological area
, the glaucoma classification subnet is developed for producing the binary labels of glaucoma, in which the multi-scale features are learned and extracted.
The introduction of multi-scale features is according to \textit{Finding 3}.

\begin{figure*}[!t]

\centering
\includegraphics[width=0.82\linewidth]{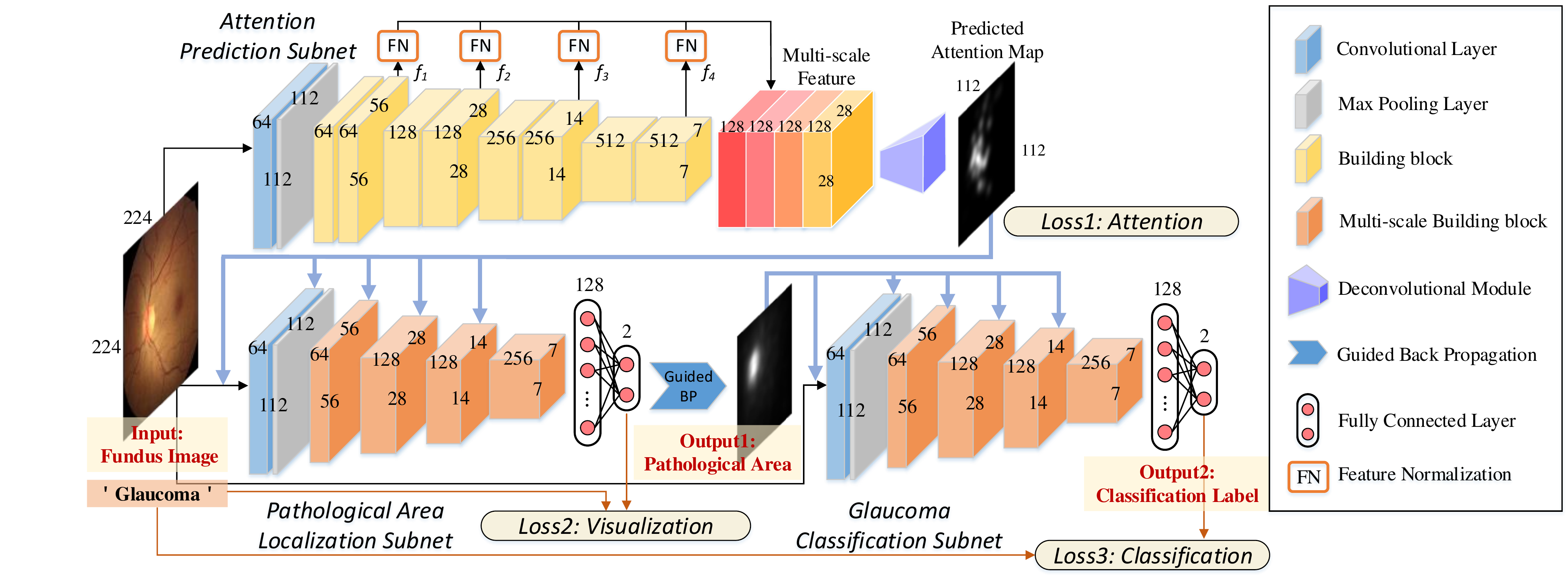}

\caption{\footnotesize Architecture of our AG-CNN network for glaucoma detection. The sizes of the feature maps and convolutional kernels are shown in this figure.
}

\label{fig:net}
\end{figure*}

The framework of AG-CNN is shown in Figure \ref{fig:net}, and its components, including multi-scale building block, deconvolutional module and feature normalization, are further demonstrated in Figure \ref{subnet}.
As shown in Figure \ref{fig:net}, the input to AG-CNN is the RGB channels of a fundus image, while the output is (1) the located pathological area and (2) the binary glaucoma label.
In addition, the located pathological area is obtained in our AG-CNN in two 2 stages. In the first stage, the ROI of glaucoma detection is learned from the attention prediction subnet, aiming to predict human attention on diagnosing glaucoma.
In the second stage, the predicted attention map is embedded in the pathological area localization subnet, and then the feature map of this subnet is visualized to locate the pathological area.
Finally, the located pathological area is further used to to mask the input and features of the glaucoma classification subnet, for outputting the binary labels of glaucoma.


The main structure of AG-CNN is based on residual networks \cite{he2016deep}, in which the basic module is building block. Note that all convolutional layers in AG-CNN are followed by a batch normalization layer and a ReLU layer for increasing the nonlinearity of AG-CNN, such that the convergence rate can be sped up. The process of training AG-CNN is in an end-to-end manner with three parts of supervision, i.e., attention prediction loss, pathological area localization loss and glaucoma classification loss.

\vspace{-0.5em}

\subsection{Attention prediction subnet}

In AG-CNN, an attention prediction subnet is designed to generate the attention maps of the fundus images, which are then used for pathological area localization and glaucoma detection. Specifically, the input of the attention prediction subnet is the RGB channels of a fundus image, which is represented by the tensor (size: $224\times224\times3$ ). Then, the input tensor is fed to one convolutional layer with kernel size of $7\times7$, followed by one max-pooling layer. Subsequently, the features flow into 8 building blocks for extracting the hierarchical features.
For more details about the building blocks, refer to \cite{he2016deep}.
Afterwards, the features of 4 hieratical building blocks are processed by feature normalization (FN), the structure of which is shown in Figure \ref{subnet} (Right).
As a result, four $28\times28\times128$ features are obtained. They are concatenated to form $28\times28\times512$ deep multi-scale features.
Given the deep multi-scale features, a deconvolutional module is applied to generate the gray attention map with the size of $112\times112\times1$.
The structure of the deconvolutional module is also shown in Figure \ref{subnet} (middle). As shown in this figure, the deconvolutional module is comprised by 4 convolutional layers and 2 deconvolutional layers.
Finally, a $112\times112\times1$ attention map can be yielded, the values of which range from $0$ to $1$.
In AG-CNN, the yielded attention maps are used to weight the input fundus images and the extracted features of the pathological area localization subnet. This is to be discussed in the next section.

\subsection{Pathological area localization subnet}

After predicting the attention maps, we further design a pathological area localization subnet to visualize the CNN feature map in glaucoma classification. The predicted attention maps can effectively make the network focus on the salient region with reduced redundancy; however, the network may inevitably miss some potential features useful for glaucoma classification. Moreover, it has been verified that the deep learning methods outperform human in the task of image classification both on nature images \cite{he2015delving,lecun2015deep} and medical images \cite{Kermany2018Identifying,Rajpurkar2017CheXNet,Poplin2017Predicting}. Therefore, we further design a subnet to visualize the CNN features for finding the pathological area.

Specifically, the pathological area localization subnet is mainly composed of convolutional layers and fully connected layers.
In addition, the predicted attention maps are used to mask the input fundus images and the extracted feature maps at different layers of the pathological area localization subnet.
The structure of this subnet is the same as the glaucoma classification subnet, which is to be discussed in section \ref{classification_subnet}.
Then, the visualization map of pathological area is yielded through guided BP \cite{springenberg2014striving} from the output of the fully connection layer to the input RGB channels fundus images.
Finally,  the visualization map is down-sampled to $112\times112$ with its values being normalized to $0-1$, as the output of the pathological area localization subnet.

\subsection{Glaucoma classification subnet}\label{classification_subnet}

In addition to the attention prediction subnet and pathological area localization subnet, we design a glaucoma classification subnet for the binary classification of positive or negative glaucoma. Similar to the attention prediction subnet, the glaucoma classification subnet is composed of one $7\times7$ convolutional layer, one max-pooling layer, 4 multi-scale building blocks.

\begin{figure}[!t]

\centering
\includegraphics[width=1.0\linewidth]{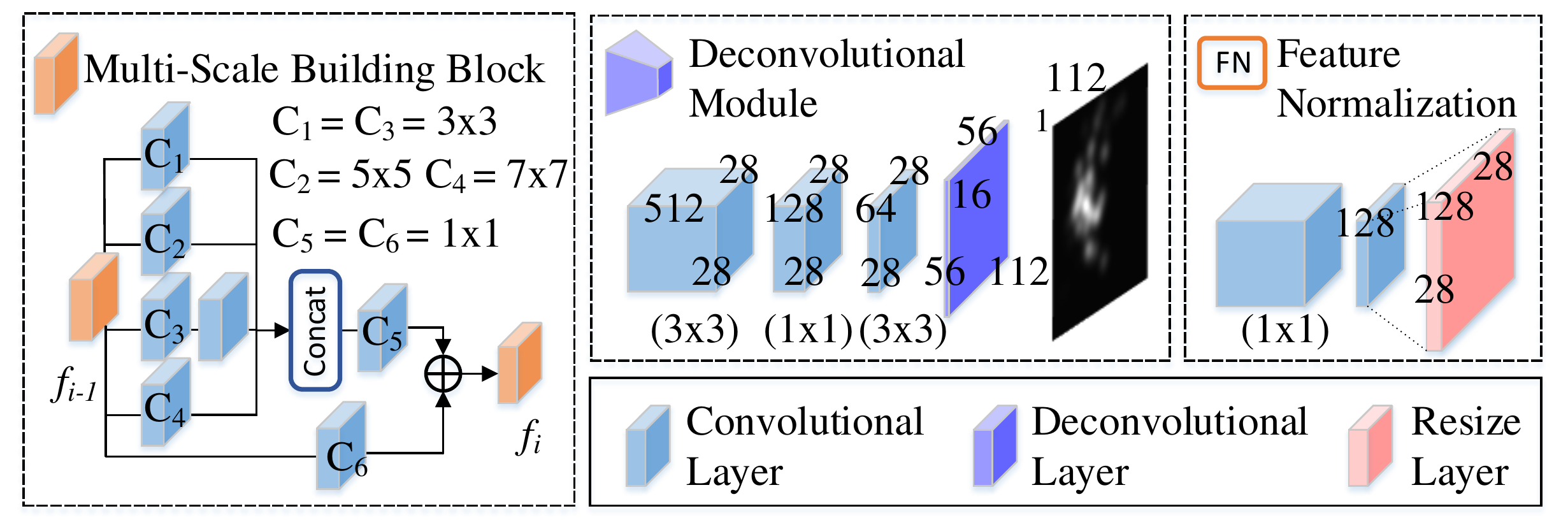}
\caption{\footnotesize Components of the AG-CNN architecture.
}

\label{subnet}

\end{figure}

The multi-scale building blocks differ from the traditional building block of \cite{he2016deep} from the following aspect. As shown in Figure \ref{subnet} (Left), 4 channels of convolutional layers $C_{1}$, $C_{2}$, $C_{3}$ and $C_{4}$ with different kernel sizes are concatenated to extract multi scale features, comparing with the traditional building block which only has a single convolutional channel. Finally, 2 fully connected layers are applied to output the classification result.

The main difference between the glaucoma classification subnet and the conventional residual network \cite{he2016deep} is that the visualization maps of pathological area weight both the input image and extracted features to focus on the ROI.
Assume that the visualization map generated by the pathological area localization subnet is $\mathbf{\hat{V}}$. Mathematically, the features $\mathbf{F}$ in the glaucoma classification subnet can be masked by $\mathbf{\hat{V}}$ as follows,
\begin{equation}
\label{eq:1}
    \mathbf{F}'=\mathbf{F}\odot\left\{(1-\theta)\mathbf{\hat{V}}\oplus\theta\right\},
\end{equation}
where $\theta$ (=0.5 in this paper) is a threshold to control the impact of the visualization map.
In the above equation, $\odot$ and $\oplus$ represent the element-wise multiplication and addition.
In the glaucoma classification subnet, the input fundus image is masked with the visualization map in the same way.
Finally, in our AG-CNN method, the redundant features irrelevant to glaucoma detection can be inhibited and the pathological area can be highlighted.

\subsection{Loss function}

In order to achieve end-to-end training, we supervise the training process of AG-CNN through attention prediction loss (denoted by  $\rm{Loss}_a$) , feature visualization loss (denoted by  $\rm{Loss}_f$)
and glaucoma classification loss (denoted by $\rm{Loss}_c$), as shown in Figure \ref{fig:net}. In our LAG database, both the glaucoma label $l$ ($\in \{0,1\}$) and the attention map $\mathbf{A}$ (with its elements $A_{i,j} \in [0,1]$) are available for each fundus image, seen as the GT in the loss function.
We assume that $\hat{l}$ ($\in \{0,1\}$)  and $\hat{\mathbf{A}}$ (with its elements $\hat{A}_{i,j} \in [0,1]$) are the predicted glaucoma label and attention map, respectively.
Following \cite{Huang2015SALICON}, we utilize the Kullback-Leibler (KL) divergence function as the human-attention loss $\rm{Loss}_a$.
Specifically, the human-attention loss is represented by

\begin{equation}
\label{loss_attention}
{\rm{Loss}}_a=\frac{1}{I\cdot J}\sum_{i=1}^I\sum_{j=1}^J A_{ij}\log(\frac{A_{ij}}{\hat{A}_{ij}}),
\end{equation}
where $I$ and $J$ are the length and width of attention maps.

Furthermore, the pathological area localization subnet and glaucoma classification subnet are all supervised by the glaucoma label $l$ based on the cross-entropy function, which measures the distance between the predicted label $\hat{l}$ and its corresponding GT label $l$.
 Mathematically, $\rm{Loss}_f$ is calculated as follows,
\begin{equation}
\label{loss_machine}
        {\rm{Loss}}_c=l\log(\frac{1}{1+e^{-\hat{l}_c}})+(1-l)\log(1-\frac{1}{1+e^{-\hat{l}_c}}),
\end{equation}
where $\hat{l}_c$ represents the predicted label from the glaucoma classification subnet.  Similar way is used to calculate Lossf, which
replaces $\hat{l}_c$ by $\hat{l}_f$ in \ref{loss_machine}.

Finally, the overall loss is the linear combination of $\rm{Loss}_a$, $\rm{Loss}_f$ and $\rm{Loss}_c$:
\begin{equation}
\label{loss_overall}
    \rm{Loss} =\alpha \cdot \rm{Loss}_a+\beta \cdot \rm{Loss}_f+\gamma \cdot \rm{Loss}_c,
\end{equation}
where $\alpha$, $\beta$ and $\gamma$ are hyper-parameters for balancing the trade-off among attention loss, visualization loss and classification loss. At the begining of training AG-CNN, we choose to set $\alpha\gg\beta=\gamma$ to speed the convergence of attention prediction subnet. Then, we set $\alpha\ll\beta=\gamma$ to minimize the feature visualization loss and the classification loss, thus realizing the convergence of prediction. Given the loss function of (\ref{loss_overall}), our AG-CNN model can be end-to-end trained for glaucoma detection and pathological location.

\begin{table}

\scriptsize
\centering
\caption{\footnotesize Performance of three methods for glaucoma detection over the test set of our LAG database.}
\label{result_LAG}

\setlength{\tabcolsep}{1mm}{
\begin{tabular}{cccccc}
    \toprule
      Method & Accuracy & Sensitivity & Specificity & AUC & $\rm{F_2\!\!-\!score}$ \\
    \midrule
      \textbf{Ours} & \textbf{95.3\%} & \textbf{95.4\%} & \textbf{95.2}\% & \textbf{0.975} & \textbf{0.951}\\
      Chen et al. & 89.2\% & 90.6\% & 88.2\% & 0.956 & 0.894 \\
      Li et al. & 89.7\% & 91.4\% & 88.4\% & 0.960 & 0.901  \\
    \bottomrule

\end{tabular}}

\end{table}
\begin{table}

\scriptsize
\centering
\caption{\footnotesize Performance of three methods for glaucoma detection over the RIM-ONE database.}
\label{result_rimone}

\setlength{\tabcolsep}{1mm}{
\begin{tabular}{cccccc}
    \toprule
      Method & Accuracy & Sensitivity & Specificity & AUC & $\rm{F_2\!\!-\!score}$ \\
    \midrule
      \textbf{Ours} & \textbf{85.2\%} & \textbf{84.8\%} & 85.5\% & \textbf{0.916} & \textbf{0.837}\\
      Chen et al. & 80.0\% & 69.6\% & \textbf{87.0}\% & 0.831 & 0.711 \\
      Li et al. & 66.1\% & 71.7\% & 62.3\% & 0.681 & 0.679  \\
    \bottomrule

\end{tabular}}

\end{table}
\section{Experiments and Results}

\subsection{Settings}
In this section, the experiment results are presented to validate the performance of our method in glaucoma detection and pathological area localization.
In our experiment, the 5,824 fundus images in our LAG database are randomly divided into training (4,792 images), validation (200 images) and test (832 images) sets. To test the generalization ability of our AG-CNN, we further validate the performance of our method on another public database RIM-ONE \cite{Fumero2011RIM}.
Before inputting to AG-CNN, the RGB channels of fundus images are all resized to $224\times224$.
In training AG-CNN, the gray attention maps are downsampled to $112\times112$ with their values normalized to be $ 0 \sim 1$.
The loss function of (\ref{loss_overall}) for training the AG-CNN model is minimized through the gradient descent algorithm with Adam optimizer \cite{Kingma2014Adam}.
The initial learning rate is $1\times10^{-5}$.
We first set $\alpha=20$ and $\beta=\gamma=1$ in (\ref{loss_overall}) until the loss of the attention prediction subnet converges, and then set $\alpha=1$ and $\beta=\gamma=10$ for focusing on the feature visualization loss and glaucoma classification loss.
Additionally, batch size is set to be 8.

Given the trained AG-CNN model, our method is evaluated and compared with two other state-of-the-art glaucoma detection methods \cite{Chen2015Glaucoma,Z2018Efficacy}, in terms of different metrics. Specifically, the metrics of sensitivity and specificity are defined as follows,
\begin{equation}\label{sensitivity}
\rm{Sensitivity}=\frac{TP}{TP+FN},
\end{equation}
\begin{equation}\label{specificity}
\rm{Specificity}=\frac{TN}{TN+FP},
\end{equation}
where $\rm{TP, TN, FP} $ and $\rm{FN}$ are the numbers of the true positive glaucoma, true negative glaucoma, false positive glaucoma and false negative glaucoma, respectively.
Based on TP, FP and FN, the $\rm{F_\beta\!\!-\!score}$ is calculated by
\begin{equation}\label{specificity}
\rm{F_\beta\!\!-\!score}=\frac{(1+\beta^2) \cdot \rm{TP}}{(1+\beta^2) \cdot \rm{TP}+\beta^2 \cdot \rm{FN} +\rm{FP}}.
\end{equation}
In the above equation, $\beta$ is the hyper-parameter balancing the trade-off between sensitivity and specificity, and it is set to $2$ as the sensitivity is more important in medical diagnosis.
In addition, receiver operating characteristic curve (ROC) and area under ROC (AUC) are also evaluated for comparing the performance of glaucoma detection.
All experiments are conducted on a computer with an Intel(R) Core(TM) i7-4770 CPU@3.40GHz, 32GB RAM and a single Nvidia GeForce GTX 1080 GPU. Benefiting from the GPU, our method is able to detect glaucoma of 30 fundus images per second, and it is comparable to 83 and 21 fundus images per second for \cite{Chen2015Glaucoma} and \cite{Z2018Efficacy}.

\begin{figure}
\centering

\includegraphics[width=.98\linewidth]{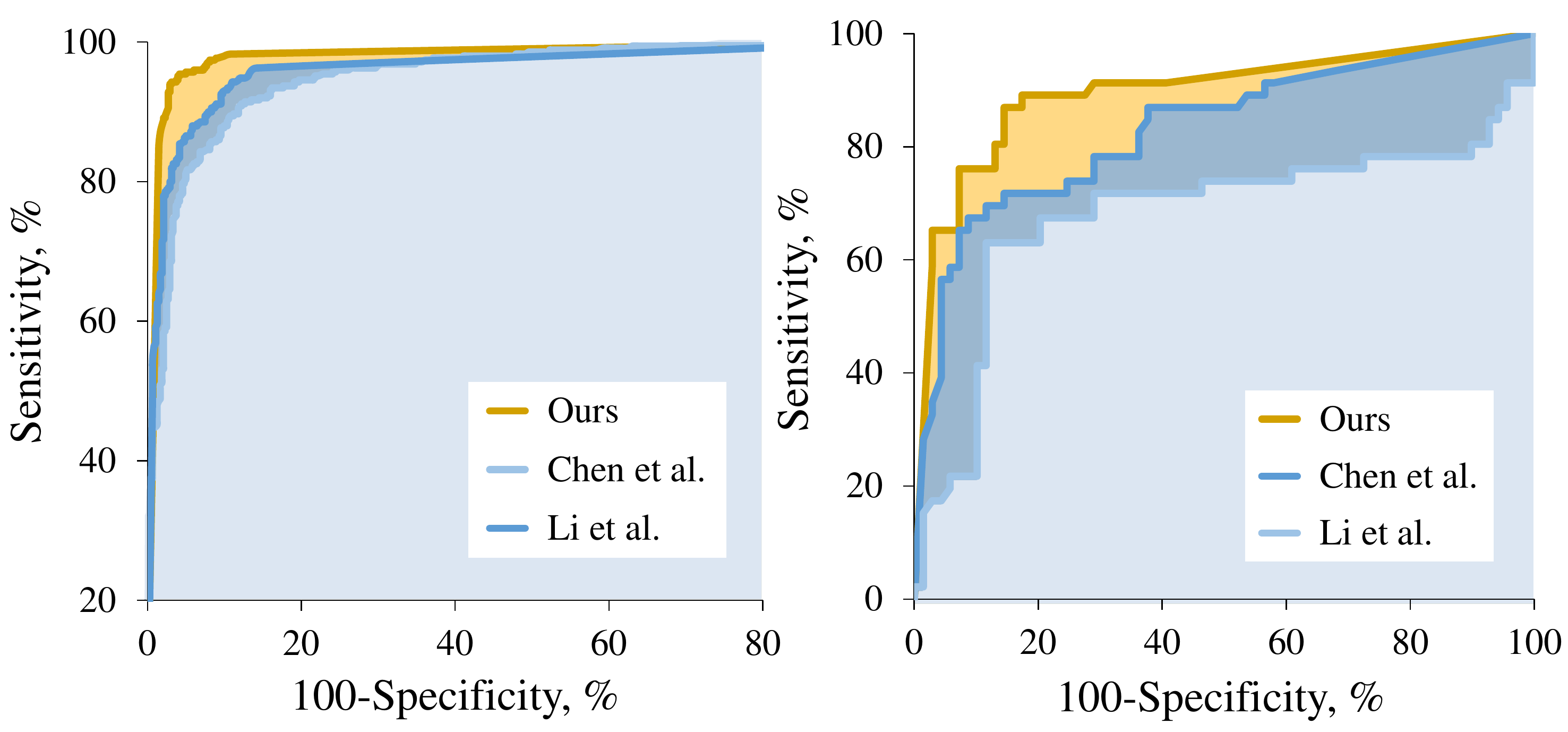}
\caption{\footnotesize Comparison of ROC curves among different methods. (Left):  Testing on our LAG testing set. (Right): Testing on RIM-ONE database.
}

\label{fig:roc}

\end{figure}

\subsection{Evaluation on glaucoma detection}

In this section, we compare the glaucoma detection performance of our AG-CNN method with two other methods \cite{Chen2015Glaucoma,Z2018Efficacy}.
Note that the models of other methods are retrained over our LAG database for fair comparison.
Table \ref{result_LAG} lists the results of accuracy, sensitivity, specificity, $\rm{F_2\!\!-\!score}$ and AUC.
As seen in Table \ref{result_LAG}, our AG-CNN method achieves 95.3\%, 95.4\% and 95.2\% in terms of accuracy, sensitivity and specificity, respectively, which are considerably better than other two methods.
Then, the $\rm{F_2\!\!-\!score}$ of our method is 0.951, while  \cite{Chen2015Glaucoma} and \cite{Z2018Efficacy} only have $\rm{F_2\!\!-\!scores}$ of 0.894 and 0.901.
The above results indicate that our AG-CNN method significantly outperforms other two methods in all metrics.

In addition, Figure \ref{fig:roc} (Left) plots the ROC curves of our and other methods, for visualizing the trade-off between sensitivity and specificity.
We can see from this figure that the ROC curve of our method is closer to the upper-left corner, when comparing with other two methods.
This means that the sensitivity of our method is always higher than those of \cite{Chen2015Glaucoma,Z2018Efficacy} at the same specificity.
We further quantify ROC performance of three methods through AUC. The AUC results are also reported in Table \ref{result_LAG}.
As shown in this table, our method has larger AUC than other two compared methods.
In summary, we can conclude that our method performs better in all metrics than \cite{Chen2015Glaucoma,Z2018Efficacy} in glaucoma detection.

To evaluate the generalization ability, we further compare the performance of glaucoma detection by our method with other 2 methods \cite{Chen2015Glaucoma,Z2018Efficacy} on the RIM-ONE database \cite{Fumero2011RIM}.
To our best knowledge, there is no other public database of fundus images for glaucoma.
The results are shown in Table \ref{result_rimone} and Figure \ref{fig:roc} (Right).
%
As shown in Table \ref{result_rimone}, all metrics of our AG-CNN method over the RIM-ONE database are above $0.83$, despite slightly smaller than the results over our LAG database. The performance of our method is considerably better than other two methods (except specificity of \cite{Z2018Efficacy}). It is worth mentioning that the metric of sensitivity is more important than that of specificity in glaucoma detection, as other indicators, e.g., intra-ocular pressure and the field of vision, can be further used for confirming the diagnosis of glaucoma.
This implies that our method has high generalization ability.

More importantly, Table \ref{result_rimone} and Figure \ref{fig:roc} (Right) show that our AG-CNN method performs significantly better than other methods especially in terms of sensitivity.
In particular, the performance of \cite{Z2018Efficacy} severely degrades, as incurring the over-fitting issue.
In a word, our AG-CNN method performs well in the generalization ability, considerably better than other state-of-the-art methods \cite{Chen2015Glaucoma,Z2018Efficacy}.




\begin{figure}[!t]

\centering
\includegraphics[width=0.6\linewidth]{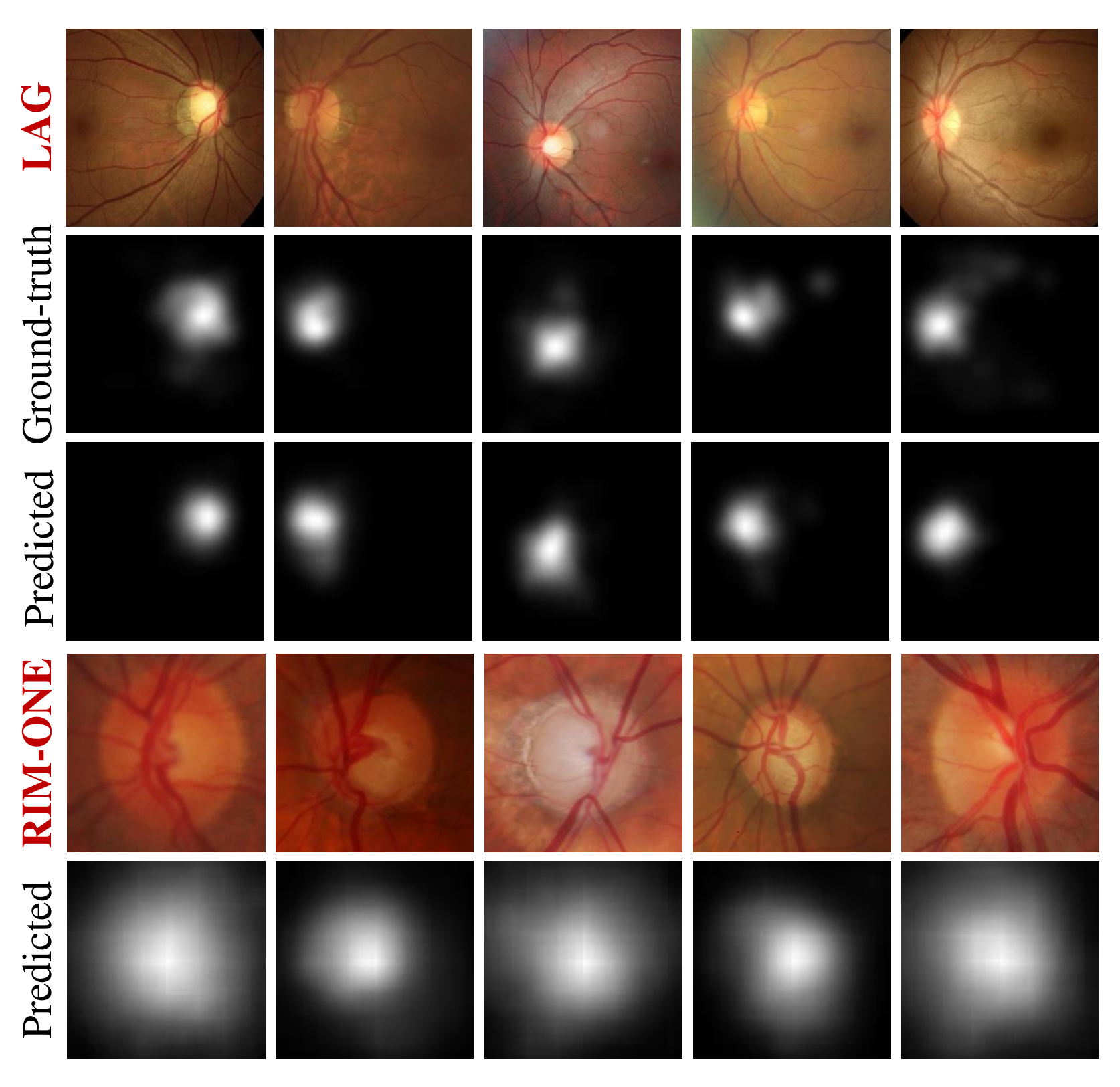}
\caption{\footnotesize Attention maps predicted by AG-CNN ramdomly selected from the test dataset. The fundus images are from our LAG (upper) and RIM-ONE (lower) database. Note that the RIM-ONE database has not the GT of the attention map.
}
\label{fig:map}

\end{figure}

\subsection{Evaluation on attention prediction and pathological area localization}

We first evaluate the accuracy of the attention model embedded in our AG-CNN model.
Figure \ref{fig:map} visualizes the attention maps predicted by our AG-CNN method over the LAG database and RIM-ONE database.
We can see from this figure that the predicted attention maps of AG-CNN are close to those of GT, when testing on our LAG database. The CC between the predicted attention maps and the GT is 0.934 on average, with a variance of 0.0032. This implies the attention prediction subnet of AG-CNN is able to predict attention maps with high accuracy.
We can further see from Figure \ref{fig:map} that the attention maps locate the salient optic cup and disc for the RIM-ONE database, in which the scales of fundus images are totally different from those of LAG database. Thus, our method is robust to the scales of fundus images in predicting attention maps.

Then, we focus on the performance of pathological area localization. Figure \ref{fig:location} visualizes the located pathological area over the LAG database. Comparing the GT pathological area with our localization results, we can see from Figure \ref{fig:location} that our AG-CNN model can accurately located the areas of optic cup and disc and the region of retinal nerve fiber layer defect, especially for the pathological areas of the upper and lower optic disc edge.

Besides, we calculate the CC between the located pathological area and the GT attention maps of ophthalmologists, with an average of 0.581 and a variance of 0.028. This also implies that (1) on one hand, the pathological area localization results are consistent with the attention maps of ophthalmologists; (2) on the other hand, the pathological area cannot be completely covered by the attention maps.
Moreover, we also compare our attention based pathological area localization results with a state-of-art method \cite{gondal2017weakly}, which is based on the CAM model \cite{zhou2015learning}. The results of \cite{gondal2017weakly} are shown in the $3^{rd}$ row of Figure \ref{fig:location}. We can see that it can roughly highlight the ROI but cannot pinpoint the tiny pathological area, e.g., the upper and lower edge of the optic disc boundary.
In some cases, \cite{gondal2017weakly} highlight the boundary of the eyeball, indicating that the CAM based methods extracted some unuseful features (i.e., redundancy) for classification.
Therefore, the pathological area localization in our approach is effective and reliable, especially compared to the CAM based method that does not incorporate human attention.

\subsection{Results of ablation experiments}

 \begin{figure}[!t]
\centering
\vspace{0em}
\includegraphics[width=.95\linewidth]{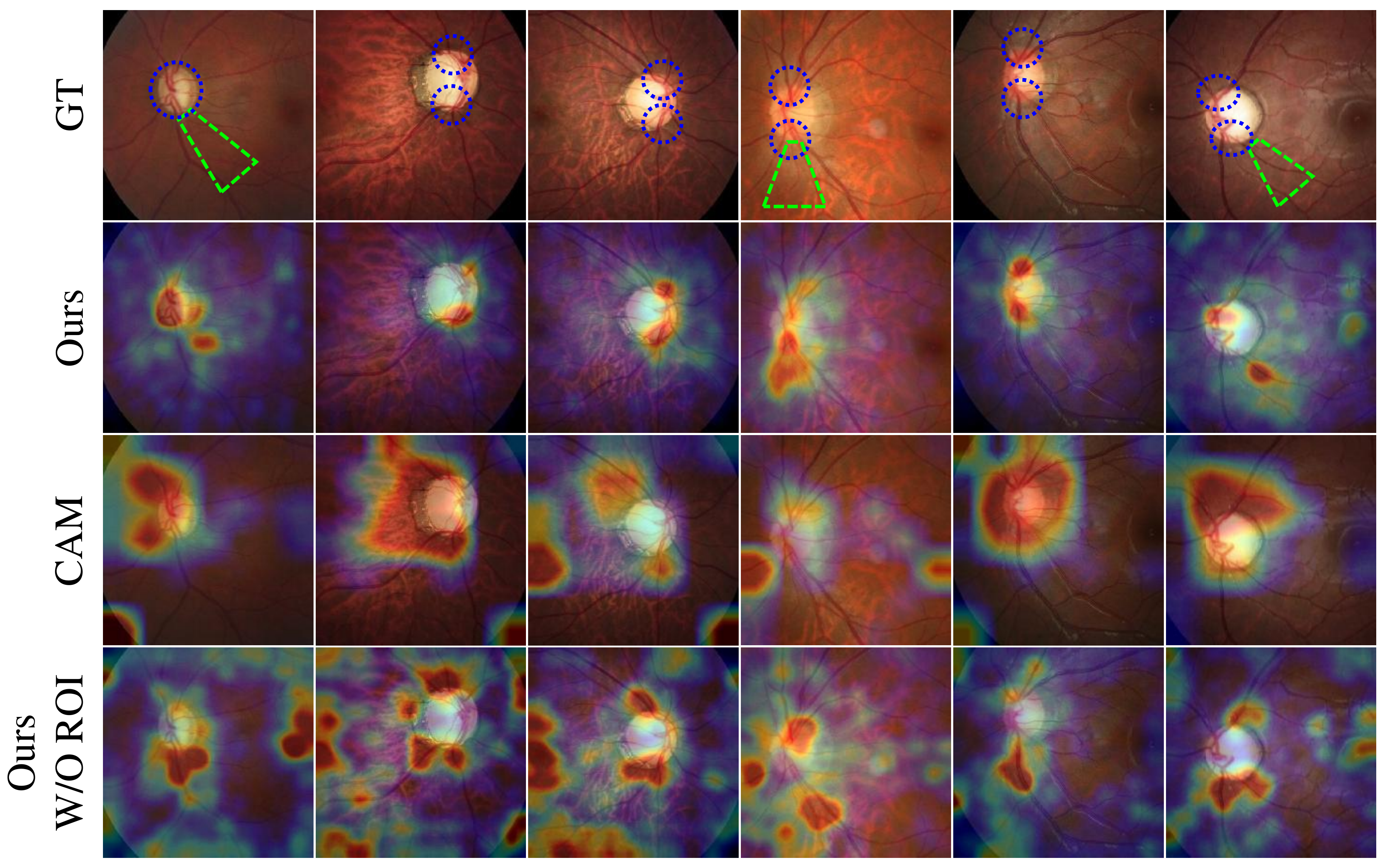}
\caption{\footnotesize Comparison of pathological area localization results for glaucoma detection. ($1^{st}$ row): The pathological areas located by ophthalmologists. Optic cup and disc are labeled in blue and the regions of retinal nerve fiber layer defect are labeled in green. ($2^{nd}$ row): The result of our method. ($3^{rd}$ row): The result of CAM based method. ($4^{th}$ row): The result of ablation experiment.
}

\label{fig:location}

\end{figure}

In our ablation experiments, we first illustrate the impact of predicted attention maps for located pathological area.
To this end, we simply remove the attention prediction subnet, and then compare the pathological localization results with and without predicted attention maps.
The results are shown in Figure \ref{fig:location}.
We can see that the pathological area can be effectively localized by using the attention maps. In contrast, the located pathological area distributes over the whole fundus image, once the attention maps are not incorporated.
Therefore, the above results verify the effectiveness and necessity of predicting the attention maps for pathological area localization in our AG-CNN approach.

Next, we assess the impact of the predicted attention map and the located pathological area on the performance of glaucoma detection. To this end, we simply remove the attention prediction subnet and pathological area localization subnet of AG-CNN, respectively, for classifying the binary labels of glaucoma. The results are shown in Table \ref{result_ablation}.
As seen in this table,
the introduction of both the predicted attention map and located pathological area can improve accuracy, sensitivity, specificity and $\rm{F_2\!\!-\!score}$ by $4.5 \%$, $4.3\%$, $4.7\%$ and $4.7\%$. However, the performance of only embedding the pathological area localization subnet and without the attention prediction subnet is even worse than removing them both. It verifies the necessity of our attention prediction subnet for pathological area localization and glaucoma detection.

Hence, the attention prediction subnet and pathological area localization subnet are able to improve the performance of glaucoma detection in AG-CNN.
Additionally, we show the effectiveness of the proposed multi-scale block in AG-CNN, via replacing it by the default conventional shortcut connection in residual network \cite{he2016deep}.
The results are also shown in  Table \ref{result_ablation}. We can see that the multi-scale block can also enhance the performance of glaucoma detection.

\begin{table}[!t]

\scriptsize
\centering
\caption{\footnotesize Ablation results over the test set of our LAG database. APS represents the attention prediction subnet. PAL represents the pathological area localization subnet.}
\label{result_ablation}

\setlength{\tabcolsep}{1mm}{
\begin{tabular}{cccccc}
    \toprule
      Method & Accuracy & Sensitivity & Specificity & AUC & $\rm{F_2\!\!-\!score}$ \\
    \midrule
      \textbf{Full AG-CNN} & \textbf{95.3\%} & \textbf{95.4\%} & \textbf{95.2\%} & \textbf{0.975} & \textbf{0.951}\\
      W APS W/O PAL&  94.0\% & 94.0\% & 94.0\% & 0.973 & 0.936 \\
      W/O APS W PAL&  87.1\% & 87.7\% & 86.7\% & 0.941 & 0.867 \\
      W/O APS W/O PAL&  90.8\% & 91.1\% & 90.5\% & 0.966 & 0.904 \\
      W/O multi-scale block&  92.2\% & 92.0\% & 92.3\% & 0.974 & 0.915 \\

    \bottomrule

\end{tabular}}

\end{table}

\section{Conclusion}
In this paper, we have proposed a new deep learning method, named AG-CNN, for automatic glaucoma detection and pathological area localization upon fundus images.
Our AG-CNN model is composed of the subnets of attention prediction, pathological area localization and glaucoma classification. As such, glaucoma could be detected using the deep features highlighted by the visualized maps of pathological areas, based on the predicted attention maps.
For training the AG-CNN model, we established the LAG database with 5,824 fundus images labeled with either positive or negative glaucoma, along with their attention maps on glaucoma detection.
The experiment results showed that the predicted attention maps significantly improve the performance of glaucoma detection and pathological area localization in our AG-CNN method, far better than other state-of-the-art methods.

\section{Acknowledgement}
This work was supported by BMSTC under Grants Z181100001918035.

{\small
\bibliographystyle{ieee_fullname}
\bibliography{ref}
}

\end{document}